\documentclass[10pt,twocolumn,letterpaper]{article}
\usepackage[pagenumbers]{iccv}
\definecolor{iccvblue}{rgb}{0.21,0.49,0.74}
\usepackage[pagebackref,breaklinks,colorlinks,allcolors=iccvblue]{hyperref}

\def\paperID{*****} % *** Enter the Paper ID here
\def\confName{ICCV}
\def\confYear{2025}

\usepackage{threeparttable}

\usepackage{epsfig}
\usepackage{graphicx}
\usepackage{amsmath,amsfonts}
\usepackage{amssymb}
\usepackage{amsthm}
\usepackage[misc]{ifsym}
\usepackage{amsfonts}
\usepackage{bm}
\usepackage{bbm}
\usepackage{array}
\usepackage{multicol}
\usepackage{multirow}
\usepackage{enumitem}
\usepackage{nccmath}
\usepackage{pifont}% http://ctan.org/pkg/pifont
\usepackage{multirow}
\usepackage{array}
\usepackage{adjustbox}
\usepackage{enumitem}
\usepackage{caption}

\usepackage{booktabs}
\usepackage[table,xcdraw,dvipsnames]{xcolor}

\definecolor{HorizonBlue}{HTML}{177cb0}

\newlength\savewidth

\newcommand{\boldparagraph}[1]{\vspace{0.2cm}\noindent{\bf #1}}

\newcolumntype{x}[1]{>{\centering\arraybackslash\hspace{0pt}}p{#1}}

\makeatletter
\def\thickhline{%
  \noalign{\ifnum0=`}\fi\hrule \@height \thickarrayrulewidth \futurelet
   \reserved@a\@xthickhline}
\def\@xthickhline{\ifx\reserved@a\thickhline
               \vskip\doublerulesep
               \vskip-\thickarrayrulewidth
             \fi
      \ifnum0=`{\fi}}
\makeatother

\newlength{\thickarrayrulewidth}
\definecolor{darkgreen}{rgb}{0.0, 0.2, 0.13}
\definecolor{darkspringgreen}{rgb}{0.09, 0.45, 0.27}

\usepackage{xspace}
\makeatletter

\newcommand{\algorithmfootnote}[2][\footnotesize]{%
  \let\old@algocf@finish\@algocf@finish% Store algorithm finish macro
  \def\@algocf@finish{\old@algocf@finish% Update finish macro to insert "footnote"
    \leavevmode\rlap{\begin{minipage}{\linewidth}
    #1#2
    \end{minipage}}%
  }%
}

\DeclareRobustCommand\onedot{\futurelet\@let@token\@onedot}
\def\@onedot{\ifx\@let@token.\else.\null\fi\xspace}

\makeatother

\begin{document}
% \title{\LaTeX\ Author Guidelines for ICCV Proceedings}
% \and
\newcommand{\methodName}{\textbf{\textit{ResAD}}}{}
\title{\methodName{}:~Normalized Residual Trajectory Modeling for \\ End-to-End Autonomous Driving}
\author{
Zhiyu Zheng$^{1,\diamond}$ \quad
Shaoyu Chen$^{2,\dagger}$ \quad
Haoran Yin$^{2}$ \quad
Xinbang Zhang$^{2}$ \quad 
Jialv Zou$^{3,\diamond}$ \\
Xinggang Wang$^{3}$ \quad 
Qian Zhang$^{2}$ \quad 
Lefei Zhang$^{1,~\textrm{\Letter}}$
\vspace{0.3em} \\
\fontsize{10.0pt}{9.84pt}
\textsuperscript{1} School of Computer Science, Wuhan University  \quad \textsuperscript{2} Horizon Robotics \\
\textsuperscript{3} School of EIC, Huazhong University of Science \& Technology \\ 
\href{https://duckyee728.github.io/ResAD}{ \ttfamily https://duckyee728.github.io/ResAD}
}
\maketitle
\let\thefootnote\relax\footnotetext{$^\diamond$ Intern of Horizon Robotics.}
\let\thefootnote\relax\footnotetext{$^\dagger$ Project lead. $^\textrm{\Letter}$ Corresponding author.}
%%%%%%%%% ABSTRACT
\begin{abstract}
End-to-end autonomous driving (E2EAD) systems, which learn to predict future trajectories directly from sensor data, are fundamentally challenged by the inherent spatio-temporal imbalance of trajectory data. 
This imbalance creates a significant optimization burden, causing models to learn spurious correlations instead of robust driving logic, while also prioritizing uncertain, distant predictions, thereby compromising immediate safety. 
To address these issues, we propose~\methodName{}, a novel \textbf{Normalized Residual Trajectory Modeling} framework. 
Instead of predicting the future trajectory directly, our approach reframes and simplifies the learning task by predicting the \textbf{residual} deviation from a deterministic \textbf{inertial reference}. 
This inertial reference serves as a strong physical prior, compelling the model to move beyond simple pattern-matching and instead focus its capacity on learning the necessary, context-driven deviations (\textit{e.g.}, traffic rules, obstacles) from this default, inertially-guided path. 
To mitigate the optimization imbalance caused by uncertain, long-term horizons, \methodName{} further incorporates \textbf{Point-wise Normalization} of the predicted residual. This technique re-weights the optimization objective, preventing large-magnitude errors associated with distant, uncertain waypoints from dominating the learning signal. 
% Extensive experiments validate the effectiveness of our framework. 
On the NAVSIM v1 and v2 benchmarks, \methodName{} achieves state-of-the-art results of 88.8 PDMS and 85.5 EPDMS with only two denoising steps, demonstrating that \methodName{} significantly simplifies the learning task and improves planning performance. 
The code will be released to facilitate further research.
\end{abstract}

\begin{figure}[t!]
    \centering
    \includegraphics[width=\columnwidth]{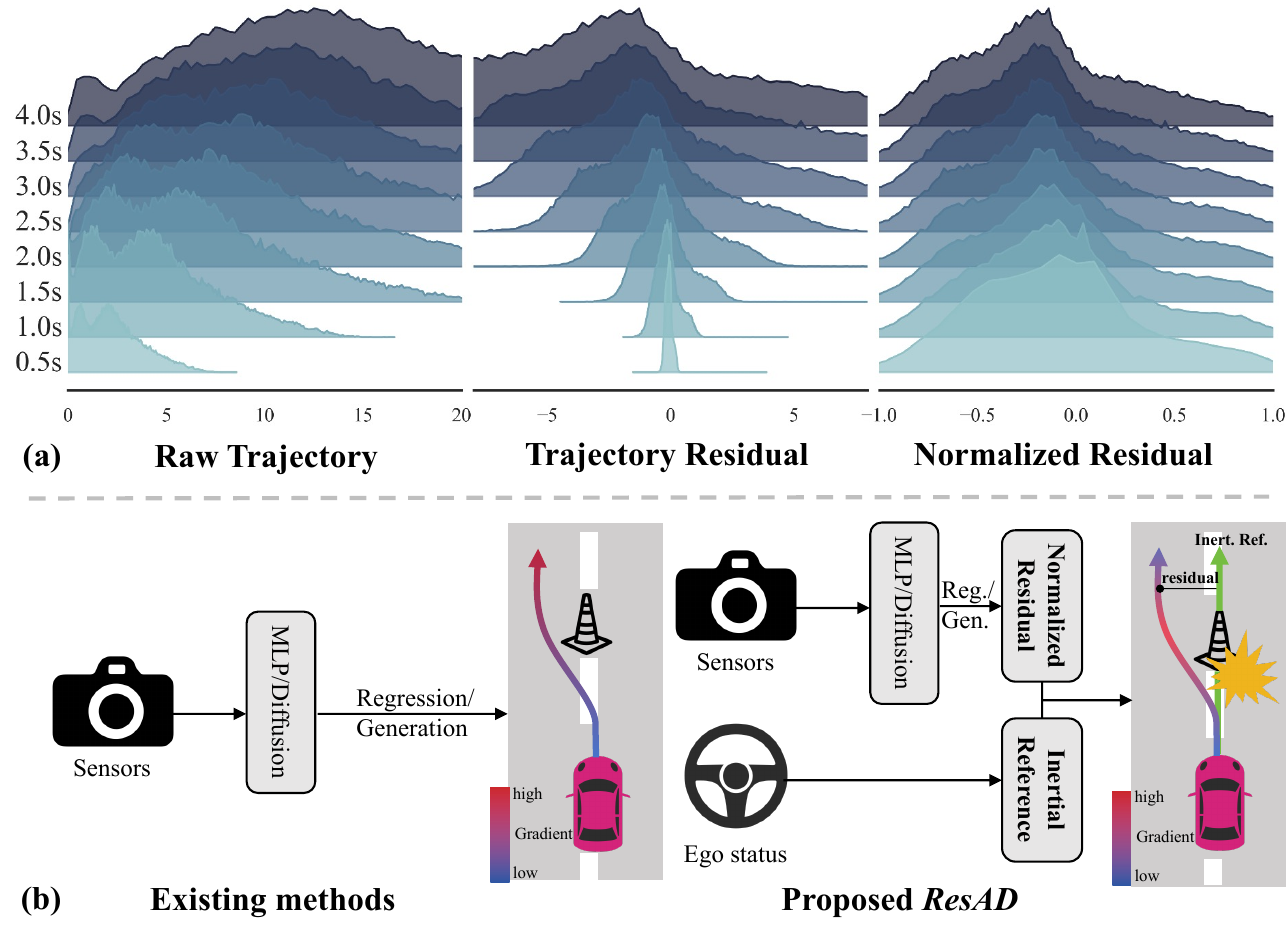}
    \caption{
    \textbf{The core motivation for~\methodName{}: Addressing the Challenge of Imbalanced Trajectory Data.}
    \textbf{(a)}~Visualization of the longitudinal data distribution of dataset trajectories under three modeling strategies. Raw Trajectories exhibit significant mean drift and increasing variance, creating a planning horizon dilemma. Our Trajectory Residual centers the distribution, and the Normalized Residual further stabilizes the variance for a simpler, balanced learning objective.
    \textbf{(b)}~Conceptual comparison. Existing methods learn complex raw trajectories directly, risking reliance on spurious correlations. Our~\methodName{} simplifies the task by learning to predict only the necessary residual deviation from a strong physical prior (the inertial reference), focusing the model on context-aware corrections.
    % \textbf{(a)}~Comparison of trajectory distributions under different modeling strategies. Raw trajectories exhibit significant mean drift and increasing variance over the prediction horizon. \textbf{Trajectory Residual Modeling} centers the distribution around zero.
    % \textbf{Point-wise Residual Normalization} further stabilizes variance for a simpler learning objective. 
    % \textbf{(b)}~Comparisons between existing methods and the proposed~\methodName{}. Instead of predicting the trajectory directly,~\methodName{} obtains an inertial reference (green arrow) as a counterfactual baseline. 
    % This simplifies the learning task: the model learns to predict the necessary deviation, rather than the entire trajectory. 
    % This effectively focuses the model on learning the mapping between scene context (like obstacles) and the required control corrections, rather than statistical correlations.
    }
    \label{fig:fig1}
\end{figure}
\section{Introduction}
Conventional autonomous driving systems rely on a modular pipeline of perception, prediction, and planning components~\cite{liang2020pnpnet,luo2018fast,chitta2022transfuser,wang2022ltp,wu2022trajectory}. This cascaded design is prone to error propagation, leading to suboptimal or unsafe driving. In response to these limitations, End-to-End Autonomous Driving (E2EAD) has emerged as a compelling alternative~\cite{hu2023uniad,chen2024end,li2024ego,sima2024drivelm}.
E2EAD reframes the driving problem by learning a direct mapping from raw sensor inputs to a future driving trajectory, from which control commands are derived, all within a single, unified framework~\cite{jia2023thinktwice,weng2024para,zhou2025opendrivevla,hamdan2025eta}.
% which remodels the driving problem by learning a direct mapping from raw sensor inputs to a future driving trajectory, from which control commands are derived, all within a single, unified framework.
% 直接预测轨迹不好，
% (1) 原始轨迹数据分布不均带来学习难题 (2) 

Recent years have witnessed extensive research into E2EAD methods, focusing on developing more effective representations~\cite{jiang2023vad,chen2024vadv2,sun2025sparsedrive,zhang2025bridging}, enhancing sensor fusion techniques~\cite{Jaeger2023ICCV,ye2023fusionad,chen2024dualat,liu2025gaussianfusion}, and designing advanced architectures~\cite{yang2024visual,jia2025drivetransformer,song2025don,RAD,zhengdiffusion}. However, these existing methods are all attempting to answer the same question: ``\textbf{\textit{What is the future trajectory?}}"
We argue this approach is inherently limited. As shown in~Fig.\ref{fig:fig1}~\textbf{(a)}, the raw trajectory data exhibits a spatio-temporal non-uniformity, which leads to two critical issues that hinder real-world reliability and safety:
% \textbf{{Spurious Correlations}} and the \textbf{{Planning Horizon Dilemma}}.
\textbf{{(1)~Spurious Correlations.}} 
The immense burden of mapping high-dimensional sensor data directly to a trajectory may cause the model to find \emph{shortcuts}, relying on spurious correlations instead of the underlying robust logic governing safe driving~\cite{jia2023driveadapter,li2024exploring,shao2023reasonnet}. For instance, a model might learn to associate braking with a lead vehicle's brake lights but fail to understand the red light causing that vehicle to stop, leading it to dangerously follow the car through an intersection.
% Mapping high-dimensional sensor data directly to driving actions is a difficult task that can push the model to learn simple tricks rather than complex reasoning. It tends to depend on misleading correlations instead of the true causes of events. For instance, a model may associate a lead vehicle's brake lights with the need to slow down, without understanding that a red light is the actual cause. This failure in reasoning could lead the model to follow the car through a red light dangerously.
% The immense burden of mapping high-dimensional sensor data directly to a trajectory incentivizes the model to find ``shortcuts," latching onto spurious correlations instead of the underlying causal logic governing safe driving. For example, a model might learn to associate braking with a lead vehicle's brake lights but fail to understand the red light causing that vehicle to stop, leading it to dangerously follow the car through an intersection.
\textbf{{(2)~Planning Horizon Dilemma.}} 
Trajectory data becomes more uncertain over longer horizons.
Consequently, predictions for these distant waypoints often diverge significantly from the eventual ground truth, leading to large loss values during training~\cite{salzmann2020trajectron++,veer2022receding,chen2024end,madjid2025trajectory,jiang2025survey}. 
This skews the optimization process, forcing the model to prioritize large, unpredictable long-term errors over the precision of the critical near-term path essential for immediate collision avoidance.

To address these challenges, we propose ~\methodName{}, a Normalized \underline{\textbf{Res}}idual Trajectory Modeling framework for End-to-End \underline{\textbf{A}}utonomous \underline{\textbf{D}}riving. 
As shown in Fig.~\ref{fig:fig1} \textbf{(b)}, our core idea is to decompose the complex prediction task into two distinct components: \textbf{(1)} a deterministic physics-based baseline, the \textbf{inertial reference}, representing the vehicle's default path in the absence of active control; and \textbf{(2)} a learned \textbf{residual}, the necessary deviations from the inertial reference. 
By focusing specifically on deviations rather than the entire trajectory, \methodName{} reframes the learning objective from ``\textbf{\textit{What is the future trajectory?}}" to ``\textbf{\textit{Why must the trajectory change?}}".% 第二个问题暂时先保留为这个，感觉还是需要修改。What is the necessary deviation from a physical prior?
This shift significantly simplifies the learning problem. By providing a strong physical prior, the model can focus its capacity on learning the mapping from sensor inputs (\textit{e.g.}, traffic rules, obstacles) to the required corrections (\textit{i.e.}, residuals), rather than expending it on learning complex spatio-temporal dynamics from scratch.
% This shift encourages the model to understand underlying causal factors (\textit{e.g.}, traffic rules, obstacles) instead of exploiting spurious correlations.
To further mitigate the adverse effects of spatial scale variations during optimization, we introduce \textbf{Point-wise Residual Normalization} for the residuals. This technique prevents high-magnitude residuals at certain trajectory points from dominating the learning signal, ensuring that numerically small yet critically important adjustments are properly captured. 
Additionally, we strategically perturb the ego-vehicle’s state, generating diverse inertial references to counteract planning errors arising from sensor inaccuracies and guide the model toward a broader spectrum of high-quality trajectories. 
% By embedding the fundamental physical prior of inertia into the model’s architecture, \methodName{} significantly simplifies the learning task, enabling more nuanced and precise driving behaviors.
In summary, our contributions are as follows:
\begin{itemize}
    \item We revisit the future-trajectory-prediction paradigm in E2EAD, and contend that the spatio-temporal non-uniformity of raw trajectory data leads to spurious correlations and the planning horizon dilemma. 
    % This encourages the proposed paradigm shift.
    % , moving from predicting the trajectory itself to modeling the reasons for its deviation.
    \item We propose~\methodName{}, an E2EAD framework based on the \textbf{Normalized Residual Trajectory Modeling}. 
    % It first obtains an \textbf{inertial reference} by extrapolating the vehicle's current state and then learns to predict the \textbf{residual}, \textit{i.e.}, the necessary deviations, relative to it. 
    It simplifies the learning task by first obtaining an \textbf{inertial reference} and then learning to predict the necessary \textbf{residual} deviations relative to it.
    We further introduce \textbf{Point-wise Residual Normalization} to mitigate optimization imbalance caused by long-horizon uncertainties.
    % We also apply Point-wise Normalization to the residuals to prevent the optimization from being dominated by long-horizon uncertainties.
    \item Extensive experiments and analyses validate the effectiveness of the proposed~\methodName{}. On the NAVSIM v1 and v2 benchmark, our method achieves state-of-the-art performance with scores of 88.8 for PDMS and 85.5 for EPDMS. 
\end{itemize}
\section{Related Work}
\subsection{End-to-End Autonomous Driving}
% --------------
% E2EAD has emerged as a promising paradigm and has witnessed significant progress~\cite{jia2023driveadapter,song2025don,yao2025drivesuprim}, aiming to overcome the inherent limitations of traditional modular pipelines, such as error accumulation and information loss between discrete components.
% UniAD~\cite{hu2023uniad} pioneered a planning-oriented design that jointly optimizes perception and forecasting for the ultimate goal of trajectory planning, effectively mitigating error propagation. 
% Similarly, VAD~\cite{jiang2023vad} introduced a fully vectorized scene representation, which streamlined the perception-to-planning pipeline and enabled the enforcement of explicit, instance-level safety constraints.
% Most recently, generative E2EAD methods have gained widespread attention~\cite{chi2023diffusion,zheng2024genad,xing2025goalflow,liao2025diffusiondrive}.
% Despite advancing E2EAD performance from various perspectives, existing methods predominantly rely on directly predicting future trajectories. 
% We rethink this prevailing approach and introduce \textbf{Normalized Residual Trajectory Modeling}. 
% The proposed method formulates trajectories by establishing an inertial reference with clear physical meaning and a learnable residual, offering a novel perspective on trajectory representation.
% --------------
End-to-end autonomous driving (E2EAD) seeks to overcome the limitations of traditional modular pipelines, such as error accumulation and inter-module information loss~\cite{jia2023driveadapter,min2024driveworld,hanselmann2022king,nie2024reason2drive,jia2024bench2drive,song2025don,sun2025generalizing}. Pioneering works like UniAD~\cite{hu2023uniad} introduced a planning-oriented architecture that jointly optimizes perception and forecasting to mitigate error propagation. VAD~\cite{jiang2023vad} further streamlined the pipeline with a fully vectorized scene representation, enabling the enforcement of explicit, instance-level safety constraints. More recently, generative models have become a new frontier in E2EAD research~\cite{chi2023diffusion,chen2024ppad,feng2025artemis,zheng2024genad,liao2025diffusiondrive,xu2025knowledge,jiang2025diffvla,li2025discrete}.
GoalFlow~\cite{xing2025goalflow} introduces a goal-conditioned generative model that first selects an optimal goal point based on scene context and then uses Flow Matching to efficiently generate high-quality trajectories towards it.
% ARTEMIS~\cite{feng2025artemis} auto-regressively generates sequential trajectories to maintain critical temporal dependencies, while dynamically dispatching scene-specific queries to expert networks.
Despite these advances, existing methods predominantly rely on the direct prediction of future trajectories. 
In this work, we depart from this paradigm by introducing \textbf{Normalized Residual Trajectory Modeling}. Our method formulates a trajectory by decomposing it into a physics-based inertial reference and a learnable residual, offering a more structured and interpretable approach to trajectory representation.
\begin{figure*}[t!]
    \centering
    \includegraphics[width=\textwidth]{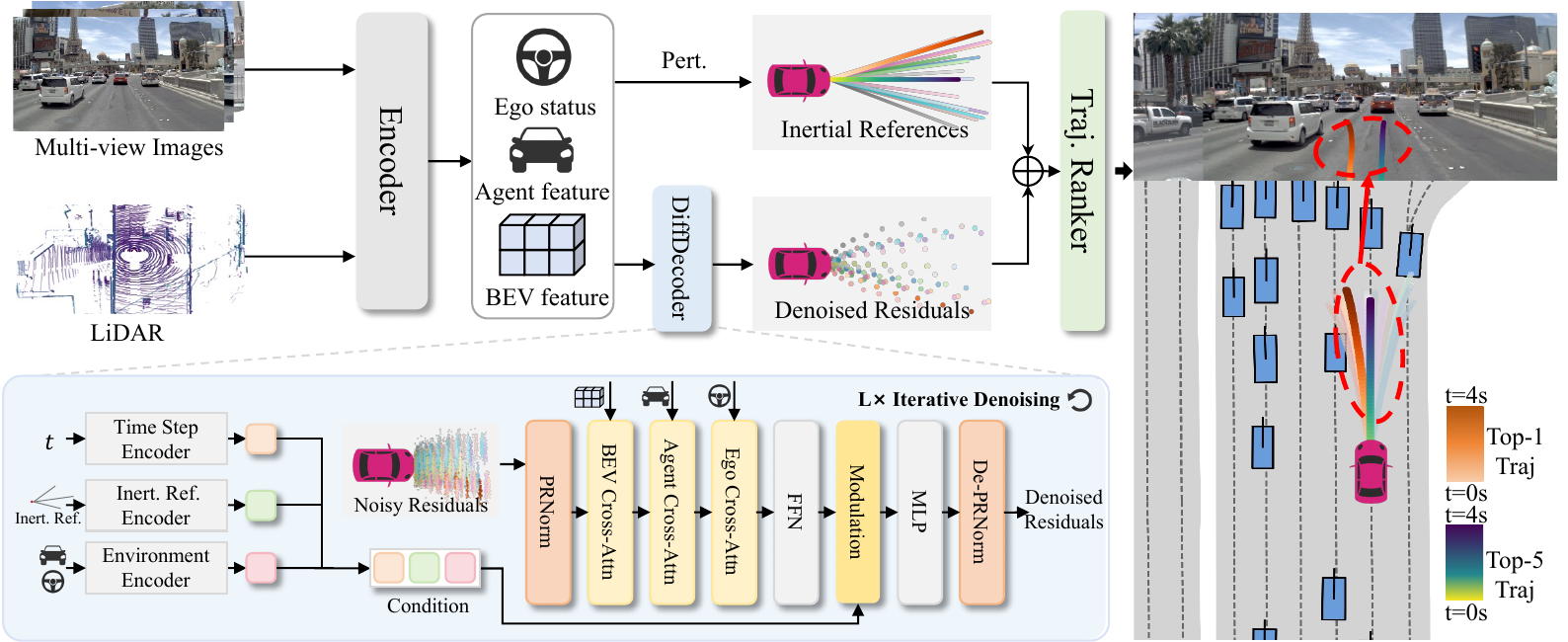}
    \caption{The proposed \methodName{} framework. Instead of predicting the entire trajectory,~\methodName{} establishes a strong physical prior based on the vehicle's current state: the Inertial Reference. By applying Inertial Reference Perturbation, the framework generates a diverse set of initial intent hypotheses. Finally, the diffusion decoder (DiffDecoder), conditioned on these references, learns to predict the necessary Normalized Residuals. We highlight the 1st-ranked and 5th-ranked output trajectories, denote as Top-1 Traj and Top-5 Traj.
    % In all figures, \textit{Top-1 Traj} refers to the 1st-ranked trajectory, and \textit{Top-5 Traj} refers to the 5th-ranked trajectory.
    % Multi-view images and LiDAR data are first processed and fused by a feature interaction encoder. We generate an inertial reference from the ego-vehicle's state and perturb it into a cluster to ensure robustness and enable multi-modal predictions. Finally, diffusion decoders, conditioned on this reference cluster, merge the encoded features via cross-attention to output the planned trajectories. 
    % Our proposed PRNorm is used to normalize the trajectory scales.
    % Multi-view images and LiDAR data are first processed and fused by a feature interaction encoder. We generate an inertial reference from the ego-vehicle's state and perturb it into a cluster to ensure robustness and enable multi-modal predictions. Finally, diffusion decoders, conditioned on this reference cluster, merge the encoded features via cross-attention to output the planned trajectories. 
    }
    \label{fig:main}
\end{figure*}

% 绝大多数端到端自动驾驶方案输出一个确定性轨迹，这可能是不现实的，由于驾驶场景的复杂性与决策的多样性，单一的输出可能导致次优结果。
\subsection{Multimodal Planning}
Most E2EAD systems produce a single, deterministic trajectory, an approach that struggles with the inherent diversity of real-world driving scenarios.
To address this, several works have explored multimodal planning~\cite{chen2024vadv2,li2025generalized,song2025breaking,guo2025ipad}.
VADv2~\cite{chen2024vadv2} proposes a probabilistic planning framework that outputs a distribution of future trajectories, which can be sampled to produce a diverse set of behaviors. 
The Hydra-MDP series~\cite{li2024hydra,li2025hydra++} employs policy distillation to select multiple trajectories from a vocabulary guided by an expert.
GTRS~\cite{li2025generalizedtrajectoryscoringendtoend} adopts a different strategy, scoring a set of pre-generated trajectories to ensure both diversity and safety.
DiffusionDrive~\cite{liao2025diffusiondrive} highlights the challenge of mode collapse in generative-based models, addressing it by anchoring trajectory generation to a fixed cluster vocabulary. 
% However, these methods are constrained by their reliance on a static, predefined vocabulary. This approach is inherently inefficient, as it evaluates numerous trajectories that are unsuitable for the current context.
However, these methods fundamentally rely on a static, predefined vocabulary.
This makes them both inefficient and restrictive, forcing them to evaluate irrelevant options while being unable to generate truly optimal trajectories outside the discrete set.
In contrast, ~\methodName{} benefits from its unique trajectory modeling strategy, which enables it to directly denoise from the Gaussian noise. This approach yields superior, context-aware multimodal trajectories without relying on a fixed vocabulary.

\section{Methodology}
\subsection{Preliminaries}
E2EAD aims to learn a unified policy, $\pi$, that directly maps raw sensor inputs, $\mathcal{O}$, to a sequence of waypoints, $\tau=\{(x_t,y_t)\}_{t=1}^{T_f}$, where $T_f$ denotes the planning horizon, and $(x_t, y_t)$ is the predicted future location of each waypoint at time $t$.
We construct the proposed~\methodName{} using a vanilla 
diffusion~\cite{ho2020denoising,songdenoising} framework. The diffusion model defines a Markovian chain of diffusion forward process $q$ by gradually
adding noise to sample data $z_0$, over a series of $T$ timesteps, which can be formulated as:
\begin{equation}q(\boldsymbol{z}_t|\boldsymbol{z}_0)=\mathcal{N}(\boldsymbol{z}_t|\sqrt{\bar{\alpha}_t}\boldsymbol{z}_0,(1-\bar{\alpha}_t)\boldsymbol{I}),
\end{equation}
where $\alpha_t = 1 -\beta_t$ and $\bar{\alpha}_t=\prod_{s=1}^t\alpha_s.$ The hyperparameters $\beta_t$ controlling the amount of noise added at each step. As $t \xrightarrow{} T$, $z_T$ approaches a pure Gaussian noise distribution. We train the denoise model $\pi_\theta$ to predict $z_0$ from $z_i$ with the guidance of conditional information $c$, where $\theta$ are the trainable parameters. At the inference stage, the trained network is used to iteratively denoise a pure noise $z_T\sim\mathcal{N}(0,\mathbf{I})$ to produce a clean data sample $z_0$, which is defined as:
\begin{equation}
p_\theta\left(z_0\mid c\right)=\int p\left(z_T\right)\prod_{i=1}^Tp_\theta\left(z_{i-1}\mid z_i,c\right)\mathrm{d}z_{1:T}.
\end{equation}
In this work, instead of directly generating the future trajectory points, we define the data samples as a set of normalized residuals.
% , i.e., $z_0 = \boldsymbol{b}$, where $\boldsymbol{b}\in\mathbb{R}^{T_f\times2}$ can be obtained from the inertial reference.
% This holistic approach can potentially mitigate compounding errors and optimize the entire driving task with a single objective. Formally, the policy seeks to map a sequence of observations to a sequence of actions.
% We conduct the~\methodName{} using a vanilla diffusion policy.

\subsection{Normalized Residual Trajectory Modeling}
As illustrated in Fig.~\ref{fig:main}, \methodName{} takes multi-view images and LiDAR point clouds as input, which are fused by a Transfuser-style encoder \cite{chitta2022transfuser}.
We first generate an inertial reference from the ego-vehicle's current state. ~\methodName{} then perturbs this reference to generate a cluster, ensuring robustness and enabling multimodal predictions. 
Finally, the Diffusion Decoders employ cross-attention to merge the encoded features, using the inertial reference cluster as a condition to guide the denoising process.
\boldparagraph{Trajectory Residual Modeling.} 
\label{sec:trm}
The core idea of~\methodName{} is to reframe the complex trajectory prediction task into a simpler, more interpretable learning problem. 
Instead of predicting the entire future trajectory from scratch, we use a constant velocity mode~\cite{rhinehart2018r2p2} to extrapolate the inertial reference trajectory from the ego-vehicle's current state, and predict the residual.
This remodeling compels the model to learn the control interventions required to deviate from the default reference, focusing on the critical, high-level driving decisions needed by the scene.
% 
% the model focuses its capacity on the critical, high-level driving decisions required by the scene, rather than wasting it on modeling simple inertial motion.
% 
% The core idea of \methodName{} is to reframe trajectory prediction as a simpler, more interpretable learning problem.
% Instead of predicting the entire future trajectory from scratch, the model focuses its capacity on the critical, high-level driving decisions required by the scene, rather than wasting it on modeling simple inertial motion.
% The inertial reference trajectory is extrapolated from the ego-vehicle's current status using a constant velocity model~\cite{rhinehart2018r2p2}. 
% This reframing compels the model to learn the control interventions required to deviate from the default path, focusing on the critical, high-level driving decisions needed by the scene.

Let the ego-vehicle's velocity be $\boldsymbol{v}_0 = (v_{x,0}, v_{y,0})$ and its position be $\mathbf{p}_0=(x_0, y_0)$ at the current time $t=0$. The inertial reference trajectory $\tau_\mathrm{ref}$ for future timesteps $t_i$ in the prediction horizon $T_f=\{t_1,t_2,...,t_N\}$ is calculated as:
\begin{equation}
\label{eq:constant}
    \mathbf{p}_{t_i}=\mathbf{p}_{0}+\mathbf{v}_{0}\cdot\Delta t_i.
    % ( x_i, y_i) = (x_0 + v_x\cdot t_i,~y_0 + v_y\cdot t_i).
\end{equation}
% This reference trajectory $\tau_\mathrm{ref}$ represents the path the vehicle would follow assuming no active control inputs.
% We then compute the difference between the human-driven ground-truth trajectory $\tau_\mathrm{gt}$, and our constructed reference trajectory $\tau_\mathrm{ref}$. This difference is the trajectory residual, $\boldsymbol{b}$:
% \begin{equation}
%     \boldsymbol{b} = \tau_\mathrm{gt}-\tau_\mathrm{ref}.
% \end{equation}
This reference $\tau_{\mathrm{ref}}$ represents the path the vehicle would follow with no control inputs. We define the trajectory residual $\boldsymbol{r}$ as the point-wise difference between the ground-truth trajectory $\tau_{\mathrm{gt}}$ and the reference trajectory $\tau_{\mathrm{ref}}$:
\begin{equation}
\boldsymbol{r} = \tau_{\mathrm{gt}} - \tau_{\mathrm{ref}}.
\end{equation}
This residual $\boldsymbol{r}$~quantifies the precise corrections a human driver applied to navigate the environment. 
The learning objective of \methodName{} is thus to predict this residual, effectively capturing the driver's decision-making process.
% The residual, $\boldsymbol{b}$, represents the active corrections the vehicle must make to the reference in order to adapt to the environment and execute its driving intent.
% \input{content/fig/Fig2}

\boldparagraph{Point-wise Residual Normalization.}
\label{sec:prnorm}
A key challenge in planning is the scale variance of coordinates across the time horizon. 
Points further in the future have numerically larger values, which can cause the optimization to be dominated by far-field errors, neglecting the fine-grained, safety-critical adjustments required in the near field. 
As shown in Fig.~\ref{fig:fig1}(a), while residual modeling mitigates this by focusing on deviations, the scale issue within the residuals themselves persists. We propose Point-wise Residual Normalization (PRNorm) to resolve this.

Given a residual trajectory $\boldsymbol{r}$, which is a sequence of $T_f$ displacement vectors $\{r_1, r_2, \dots, r_{T_f}\}$, where each $r_t = (r_{t}^x,r_{t}^y)$ is a 2D vector.
% ground-truth or predicted trajectory residual as a sequence of $T_f$ vectors, $\boldsymbol{b}=\{r_1, r_2, \dots, r_{T_f}\}$.
% Each vector $r_t$ at timestep $t$ is a D-dimensional vector, $r_t = [c_t^1,c_t^2,\ldots,c_t^D]$, where each $c_t^D$ represents a spatial residual, in our setting D=2, \textit{i.e.}, $\boldsymbol{r}_{t}=[r_{t}^x,r_{t}^y]$.
A standard min-max scaling is performed on a component-wise basis for each dimension $d \in \{x,y\}$.
The extremal values, $r_{\min}^d$ and $r_{\max}^d$ are pre-computed across all timesteps and all trajectories in the entire training dataset:
% \begin{equation}
% c_{\min}^d =\min_{t\in\{1,\ldots,M\}}(c_t^d),~
% c_{\max}^d =\max_{t\in\{1,\ldots,M\}}(c_t^d).
% \end{equation}
\begin{equation}
r_{\min}^d = \min_{j, t} (r_{j,t}^d), \quad
r_{\max}^d = \max_{j, t} (r_{j,t}^d),
\end{equation}
where $j$ indexes trajectories in the training set and $t$ indexes the timestep. These values define the tightest axis-aligned bounding box for the residual. To provide fine-grained control over the final feature distribution, we introduce a hyperparameter $\gamma > 0$. 
This parameter defines the bounds of the symmetric output interval $[-\gamma,\gamma]$. The complete transformation of PRNorm for each component $r_t^d$ of every vector $r_t$ is given by:
% \begin{equation}
% \hat{c}_{t}^{d}=2\alpha\left(\frac{c_{t}^{d}-c_{\mathrm{min}}^{d}}{c_{\mathrm{max}}^{d}-c_{\mathrm{min}}^{d}+\epsilon_0}\right)-\alpha.
% \end{equation}
\begin{equation}
\tilde{r}_{t}^{d} = 2\gamma\left(\frac{r_{t}^{d}-r_{\mathrm{min}}^{d}}{r_{\mathrm{max}}^{d}-r_{\mathrm{min}}^{d}+\epsilon_0}\right) - \gamma.
\end{equation}
The small constant $\epsilon_0$ is added to the denominator to ensure numerical stability. Through this, we can get the normalized residual $\tilde{\boldsymbol{r}} = \text{PRNorm}(\boldsymbol{r})$.

% 以往的方法中，预测的未来轨迹点的尺度范围会随着速度和驾驶场景的变化而剧烈波动，这可能导致模型在优化过程中被远端点主导，而忽略了对近端点的精细调整，而近端点往往对安全性更加重要。尽管我们通过预测相对于惯性参考的残差的方式使得模型更加关注相对于参考变化更大的区域，但残差的数据大小随远近端点的尺度变化仍然存在。

\begin{table*}[t!]
  \centering
  \caption{\textbf{Performance on the NAVSIM v1 \textsc{navtest} Benchmark.}}
  \label{tab:navsim_v1}
 \begin{tabular}{lc|c|ccccc>{\columncolor{gray!20}}c}
  \toprule
  Method & Input & Backbone & \textbf{NC}~$\uparrow$ & \textbf{DAC}~$\uparrow$ & \textbf{EP}~$\uparrow$ & \textbf{TTC}~$\uparrow$ & \textbf{C}~$\uparrow$ & \textbf{PDMS}~$\uparrow$ \\
  \midrule
  % ADMLP~\cite{zhai2023rethinking} & - & Resnet-34 & 93.0 & 77.3 & 62.8 & 83.6 & \textbf{100} & 65.6 \\
  % LTF~\cite{chitta2022transfuser} & C & Resnet-34 & 97.4 & 92.8 & 79.0 & 92.4 & \textbf{100} & 83.8 \\
  Transfuser~\cite{chitta2022transfuser}~\textcolor{gray}{\scriptsize{[TPAMI'22]}} & C \& L & Resnet-34 & 97.7 & 92.8 & 79.2 & 92.8 & \textbf{100} & 84.0 \\
  UniAD~\cite{hu2023uniad}~\textcolor{gray}{\scriptsize{[CVPR'23]}} & C & Resnet-34 & 97.8 & 91.9 & 78.8 & 92.9 & \textbf{100} & 83.4 \\
  VADv2~\cite{chen2024vadv2}~\textcolor{gray}{\scriptsize{[Arxiv'24]}} & C \& L & Resnet-34 & 97.2 & 89.1 & 76.0 & 91.6 & \textbf{100} & 80.9 \\
  PARA-Drive~\cite{weng2024para}~\textcolor{gray}{\scriptsize{[CVPR'24]}} & C & Resnet-34 & 97.9 & 92.4 & 79.3 & 93.0 & 99.8 & 84.0 \\
  DRAMA~\cite{yuan2024drama}~\textcolor{gray}{\scriptsize{[Arxiv'24]}} & C \& L & Resnet-34 & 98.0 & 93.1 & 80.1 & \underline{94.8} & \textbf{100} & 85.5 \\
  Hydra-MDP~\cite{li2024hydra}~\textcolor{gray}{\scriptsize{[Arxiv'24]}}  & C \& L & Resnet-34 & \underline{98.3} & 96.0 & 78.7 & 94.6 & \textbf{100} & 86.5 \\
  Hydra-MDP++~\cite{li2025hydra++}~\textcolor{gray}{\scriptsize{[Arxiv'25]}} & C & Resnet-34 & 97.6 & 96.0 & 80.4 & 93.1 & \textbf{100} & 86.6 \\
  ARTEMIS~\cite{feng2025artemis}~\textcolor{gray}{\scriptsize{[Arxiv'25]}} & C \& L & Resnet-34 & \underline{98.3} & 95.1 & 81.4 & 94.3 & \textbf{100} & 87.0 \\
  GoalFlow~\cite{xing2025goalflow}~\textcolor{gray}{\scriptsize{[CVPR'25]}}  & C \& L & Resnet-34 & \underline{98.3} & 93.8 & 79.8 & 94.3 & \textbf{100} & 85.7 \\
  DiffusionDrive~\cite{liao2025diffusiondrive}~\textcolor{gray}{\scriptsize{[CVPR'25]}}  & C \& L & Resnet-34 & {98.2} & 96.2 & \underline{82.2} & 94.7 & \textbf{100} & 88.1 \\
  DistillDrive~\cite{yu2025distilldrive}~\textcolor{gray}{\scriptsize{[ICCV'25]}} & C \& L & Resnet-34 & {98.1} & {94.6} & 81.0 & {93.6} & \textbf{100} & {86.2} \\

  WoTE~\cite{li2025end}~\textcolor{gray}{\scriptsize{[ICCV'25]}} & C \& L & Resnet-34 & \textbf{98.5} & \underline{96.8} & 81.9 & \textbf{94.9} & \underline{99.9} & \underline{88.3} \\
  ~\methodName{}~(Ours) & C \& L & Resnet-34 & 98.0 & \textbf{97.5} & \textbf{83.3} & 94.1 & \textbf{100} & \textbf{88.8} \\
  % \bottomrule 
  % \toprule 
  \midrule
    Hydra-MDP~\cite{li2024hydra}~\textcolor{gray}{\scriptsize{[Arxiv'24]}} & C \& L & V2-99 & 98.0 & \underline{97.8} & \underline{86.5} & 93.9 & \textbf{100} & 90.3 \\
    GoalFlow~\cite{xing2025goalflow}~\textcolor{gray}{\scriptsize{[CVPR'25]}} & C \& L & V2-99 & \underline{98.4} & \textbf{98.3} & 85.0 & \underline{94.6} & \textbf{100} & 90.3 \\

  % \midrule
~\methodName{}~(Ours) & C \& L & V2-99 &  \textbf{98.9} &  \underline{97.8}  &\textbf{87.0} & \textbf{94.9}  &\textbf{100}  & \textbf{90.6} \\

  % 88.8
  \bottomrule
 \end{tabular}
 \vspace{0.2cm}
 % 这里把脚注写到表格标题里
  
  \raggedright \footnotesize ``C'': Camera, ``L'': LiDAR. The best and second-best scores are highlighted in \textbf{bold} and \underline{underlined}, respectively. 
  % * For fair comparison, we use the official scores of versions with the same backbone.
\end{table*}

\boldparagraph{Inertial Reference Perturbation.}
\label{sec:irp}
% 学习从错误的参考中“纠偏”
Driving is an inherently multi-modal task. 
Most methods depend on a fixed trajectory vocabulary, where most options are irrelevant to the current scene, causing inefficiency.
\methodName{}~generates multimodal trajectories by perturbing Inertial Reference. 
This approach offers a dual benefit.
First, it generates diverse intent hypotheses by creating varied inertial references and corresponding residuals. Second, it compels the model to handle noise inherent in ego-sensors such as GPS and IMU.

Specifically, we introduce stochastic perturbations directly into the initial velocity $\boldsymbol{v}_0$. We generate $K$ distinct perturbation vectors $\boldsymbol{\delta}_{\mathbf{v},k}$ by sampling from a zero-mean multivariate Gaussian distribution:
\begin{equation}
\boldsymbol{\delta}_{\mathbf{v},k} \sim \mathcal{N}(\mathbf{0}, \boldsymbol{\Sigma}) \quad \text{for} \quad k=1, \ldots, K.
\end{equation}
Here, the covariance matrix $\boldsymbol\Sigma=\mathrm{diag}(\sigma_{vx}^{2},\sigma_{vy}^{2})$ governs the variance of the perturbations along the longitudinal and lateral axes, respectively. These hyperparameters effectively define the exploration scope of our model's initial hypotheses. Each perturbation is additively fused with the original velocity vector to forge $K$ unique, perturbed initial states. 
By propagating each of these perturbed velocity vectors $\mathbf{v}_{0,k}^{\prime}$ through the constant velocity model Eq.~\ref{eq:constant}, we generate a set of $K$ distinct inertial references $\{\tau_{\mathrm{ref},~k}\}_{k=1}^{K}$ and corresponding residuals: 
\begin{equation}
\begin{aligned}
\mathbf{v}_{0,k}^{\prime}&=\mathbf{v}_{0}+\mathbf{\delta}_{\mathbf{v},k},\\
\{\boldsymbol{r}_k\}_{k=1}^{K}&=\{\tau_\mathrm{gt} - \tau_{\mathrm{ref},~k}\}_{k=1}^{K}.
% \mathbf{b}_{0,k}^{\prime}&=\mathbf{v}_{0}+\mathbf{\delta}_{\mathbf{v},k}.
\end{aligned}
\end{equation}
% This strategy offers two distinct advantages. Primarily, it forces the model to learn resilience against noise from ego-sensors like GPS and IMU. Furthermore, the cluster of perturbed inertial references, which encapsulates a spectrum of minor variations, acts as a set of plausible "intent hypotheses." The network is then conditioned on each hypothesis to produce a unique, complete trajectory, resulting in a naturally formed, context-relevant set of multi-modal predictions.
% 1）让网络学会对抗自车状态的噪声   2）促进多模态特性
% Inertia Reference perturbation

\boldparagraph{Training and Inference.}
In training, adding Gaussian noise to the residual cluster normalized by PRNorm:
\begin{equation}
\boldsymbol{z}{_k^{(i)}}=\sqrt{\bar{\alpha}_i}\tilde{\boldsymbol{r}}_k+\sqrt{1-\bar{\alpha}_i}\boldsymbol{\epsilon},\quad\boldsymbol{\epsilon}\sim\mathcal{N}(0,\mathbf{I}),
\end{equation}
where $\tilde{\boldsymbol{r}}_k= \mathrm{PRNorm}(\boldsymbol{r}_k )$. The diffusion decoder $f_\theta$ takes $K$ noisy normed trajectory residuals to generate denoised residuals $ \{\hat{\boldsymbol{r}}_k\}_{k=1}^K$:
\begin{equation}
\{\hat{\boldsymbol{r}}_k\}_{k=1}^K = f_\theta(\{\boldsymbol{z}^{(i)}_k\}_{k=1}^K, c),
\end{equation}
% \begin{equation}
% {\tilde{\boldsymbol{r}}_k}_{k=1}^K = f_\theta({\mathbf{z}_k^{(i)}}_{k=1}^K, c, i)
% \end{equation}
where $c$ represents the conditional information. Note that $c$ is composed of query features extracted from the encoder and the corresponding timestep embedding. Crucially, $c$ also incorporates unique positional encoding features derived from each of the perturbed inertial references. 
These encodings are essential for the model to distinguish between the different intent hypotheses and subsequently generate a diverse set of trajectories. The diffusion loss is computed as:
\begin{equation}
\mathcal{L}_\mathrm{diff}=\sum_{k=1}^{K}\mathcal{L_\mathrm{rec}}(\hat{\boldsymbol{r}}_k,\boldsymbol{r}_{k}),
\end{equation}
here, $\mathcal{L}_\mathrm{rec}$ can be a simple L1 loss or MSE loss. 

During inference, the denoising process starts with $K_\mathrm{infer}$ Gaussian noise to generate residuals. We take 2 timesteps to get the final predictions $\{\hat{\boldsymbol{r}}_k\}_{k=1}^{K_\mathrm{infer}}$ by DDIM~\cite{songdenoising}.
Then the predicted residuals are added to the corresponding perturbed inertial reference to get the multi-modal trajectory $\{\hat{\tau}_k\}_{k=1}^{K_\mathrm{infer}}$.
% \begin{equation}
% \hat\tau_k = \tau_{\mathrm{ref, k}}+\mathrm{PRNorm^{-1}(\tilde{\boldsymbol{r}}_k)}
% \end{equation}

\subsection{Multimodal Trajectory Ranker}% ranker
Inspired by VADv2~\cite{chen2024vadv2} and Hydra-MDP~\cite{li2024hydra}, we develop a Trajectory Ranker to select the optimal trajectory from multiple candidates generated by the planning module.
Given a set of trajectory candidates $v_k$, where k is the number of trajectories, we feed them into a Transformer to facilitate interaction with the perception representations, $E_{env}$, which can be expressed as follows:
% our process begins by clustering ground-truth trajectories to form a trajectory vocabulary, $v_k$, where k is the vocabulary size. 
% $v_k$ is then concatenated with the multi-modal outputs, $v_m$, from the planning head. 
% The resulting tensors serve as inputs to a Transformer, which facilitates interaction with the environmental representations, $E_{env}$, which can be expressed as follows:
\begin{equation}\begin{aligned}
\mathcal{V}&=\mathrm{PosEmb}(v_k), \\
\mathcal{V}^{\prime}&=\mathrm{Transformer}(Q=\mathcal{V},K,V=E_{env})+E.
\end{aligned}\end{equation}
$\mathrm{PosEmb}(\cdot)$ denotes the position embedding, and ego status $E$ is embedded into the transformer output. 
Subsequently, the latent vector $\mathcal{V}^{\prime}$ is fed into a set of MLP heads to predict the score $\{\hat{\mathcal{S}}_i^m|i=1,...,k\}_{m=1}^{|M|}$ for each metric $m \in M$ and the $i$-th trajectory, where M represents the set of metrics used in PDMS or EPDMS.
The ranker is trained with the ground truth score $\{{\mathcal{S}}_i^m|i=1,...,k\}_{m=1}^{|M|}$ to distill the knowledge from the rule-based planner and the ground truth waypoints as follows:
\begin{equation}
    \mathcal{L}_\mathrm{ranker} = \sum_{i=1}^ky_i\log(\hat{\mathcal{S}}_i^{im}) + \sum_{m,i}\text{BCE}({S}_i^m, \hat{\mathcal{S}}_i^m),
\end{equation}
where $y_i=\frac{e^{-({\tau}_{\mathrm{gt}}-\hat\tau_i)^2}}{\sum_{j=1}^ke^{-({\tau}_\mathrm{gt}-\hat\tau_j)^2}}.$ 
During inference, we compute scores for the outputs of the planning head and select the trajectory with the highest weighted score as the final output.
% \begin{equation}
%   \mathcal{S}_i^m,~\mathcal{S}_i^{im} = \text{MLP}^m (\mathcal{V}),~\text{MLP}^{im} (\mathcal{V}),
% \end{equation}
% \begin{equation}
%     \mathcal{L}_{ranker} = \sum_{i=1}^ky_i\log(\mathcal{S}_i^{im}) + \sum_{m,i}\text{BCE}(\hat{\mathcal{S}}_i^m, {S}_i^m).
% \end{equation}
% \begin{equation}
%     \mathcal{L}_{im} = \sum_{i=1}^ky_i\log(\mathcal{S}_i^{im}),
% \end{equation}
% \begin{equation}
%     \mathcal{L}_{kd} = \sum_{m,i}\text{BCE}(\hat{\mathcal{S}}_i^m, {S}_i^m)
% \end{equation}
% \begin{equation}\begin{aligned}
% \mathcal{L}_{ranker}=&\sum_{m,i}\hat{\mathcal{S}}_i^m\log\mathcal{S}_i^m+(1-\hat{\mathcal{S}}_i^m)\log(1-\mathcal{S}_i^m)\\
% +&\sum_{i=1}^ky_i\log(\mathcal{S}_i^{im}),
% \end{aligned}\end{equation}
% where $y_i=\frac{e^{-(\hat{T}-T_i)^2}}{\sum_{j=1}^ke^{-(\hat{T}-T_j)^2}}.$

 % & p(a)=\mathrm{MLP}(\text{Transformer}(E(a),E_{\mathrm{env}})+E_{\mathrm{navi}}+E_{\mathrm{state}}), \\
 % & q=E(a),k=v=E_{\mathrm{env}}, \\
 % & a=(x_{1},y_{1},x_{2},y_{2},...,x_{\mathrm{T}},y_{\mathrm{T}}), \\
 % & E(a)=\mathrm{Cat}\left[\Gamma(x_1),\Gamma(y_1),\Gamma(x_2),\Gamma(y_2),...,\Gamma(x_\mathrm{T}),\Gamma(y_\mathrm{T})\right], \\
 % & \Gamma(pos)=\mathrm{Cat}\left[\gamma(pos,0),\gamma(pos,1),...,\gamma(pos,L-1)\right], \\
 % & \gamma(pos,j)=\mathrm{Cat}\left[\cos(pos/10000^{2\pi j/L},\sin(pos/10000^{2\pi j/L}\right].

\section{Experiments}

\begin{table*}[h!]
\centering
\caption{\textbf{Performance on the NAVSIM v2 \textsc{navtest} Benchmark with Extended Metrics.}}
\label{tab:navsimv2}
% \resizebox{\linewidth}{!}{
\begin{tabular}{l|ccccccccc>{\columncolor{gray!20}}c}
\toprule
{Method} & \textbf{NC $\uparrow$} & \textbf{DAC $\uparrow$} & \textbf{DDC $\uparrow$} & \textbf{TL $\uparrow$} & \textbf{EP $\uparrow$} & \textbf{TTC $\uparrow$} & \textbf{LK $\uparrow$} & \textbf{HC $\uparrow$} & \textbf{EC $\uparrow$} & \textbf{EPDMS $\uparrow$} \\
\midrule
% Human Agent& 100 & 100 & 99.8 & 100 & 87.4 & 100 & 100 & 98.1 & 90.1 & 90.3 \
Ego Status MLP & 93.1 & 77.9 & 92.7 & 99.6 & 86.0 & 91.5 & 89.4 & 98.3 & 85.4 & 64.0 \\
Transfuser~\cite{chitta2022transfuser}~\textcolor{gray}{\scriptsize{[TPAMI'22]}} & 96.9 & 89.9 & 97.8 & \underline{99.7} & 87.1 & 95.4 & 92.7 & 98.3 & 87.2 & 76.7 \\
HydraMDP++~\cite{li2025hydra++}~\textcolor{gray}{\scriptsize{[Arixv'25]}} & 97.2 & \textbf{97.5} & \underline{99.4} & 99.6 & 83.1 & 96.5 & 94.4 & {98.2} & 70.9 & 81.4 \\
DriveSuprim~\cite{yao2025drivesuprim}~\textcolor{gray}{\scriptsize{[Arixv'25]}} & {97.5} & 96.5 & \underline{99.4} & 99.6 & \textbf{88.4} & 96.6 & 95.5 & 98.3 & 77.0 & {83.1} \\
ARTEMIS~\cite{feng2025artemis}~\textcolor{gray}{\scriptsize{[Arixv'25]}} & \textbf{98.3} & {95.1} &{98.6} &\textbf{99.8} &81.5 &\textbf{97.4} &{96.5} & 98.3 &- &{83.1} \\
DiffusionDrive~\cite{liao2025diffusiondrive}~\textcolor{gray}{\scriptsize{[CVPR'25]}} &\underline{98.2} &95.9 &\underline{99.4}&\textbf{99.8} &{87.5} & \underline{97.3} & \underline{96.8} & 98.3 & \underline{87.7} & \underline{84.5}\\
\midrule
~\methodName{}(Ours) &97.8 & \underline{97.2} & \textbf{99.5} & \textbf{99.8} & \underline{88.2} & 96.9 &\textbf{97.0} & \textbf{98.4} &\textbf{88.2} &\textbf{85.5} \\
\bottomrule
\end{tabular}
% }
\end{table*}
\subsection{Benchmark}
% 我们在Navsim数据集上对所提出的方法进行了验证。
We evaluate the proposed~\methodName{} on the NAVSIM v1~\cite{dauner2024navsim} and NAVSIM v2~\cite{Cao2025ARXIV} benchmark. 
NAVSIM is built upon the real-world NuPlan dataset~\cite{karnchanachari2024towards} and exclusively features relevant annotations and sensor data sampled at 2 Hz. 
The NAVSIM dataset contains two parts: \textsc{navtrain} and \textsc{navtest}, including 1192 and 136 scenarios respectively, used for trainval and test.

\boldparagraph{NAVSIM v1.} In this benchmark, each predicted trajectory is sent to a simulator, which validates the driving metrics in the corresponding environment. The planning capabilities of models are assessed using the PDM score (PDMS), which is calculated as follows:
\begin{equation}
\mathrm{PDMS}=\mathrm{NC}\times \mathrm{DAC}\times\frac{(5\times \mathrm{TTC}+2\times \mathrm{C}+5\times \mathrm{EP})}{12},
\end{equation}
where the sub-metrics NC, DAC, TTC, C, EP represent the No At-Fault Collisions, Drivable Area Compliance, Time to Collision, Comfort, and Ego Progress.

\boldparagraph{NAVSIM v2.} A new Extended PDM Score (EPDMS) is introduced in NAVSIM v2, which can be formulated as:
\begin{multline}
\mathrm{EPDMS} = \mathrm{NC} \times \mathrm{DAC} \times \mathrm{DDC} \times \mathrm{TL} \times \\
\frac{(5 \times \mathrm{TTC} + 2 \times \mathrm{C} + 5 \times \mathrm{EP} + 5 \times \mathrm{LK} + 5 \times \mathrm{EC})}{22}.
\end{multline}
The extended sub-metrics DDC, TL, LK, and EC correspond to the Driving Direction Compliance, Traffic Lights Compliance, Lane Keeping Ability, and Extended Comfort.
\subsection{Implementation Details}
For fair comparison, our model adopts an identical perception module and ResNet-34~\cite{resnet} or V2-99~\cite{v299} as its backbone following existing methods~\cite {chitta2022transfuser,li2025hydra++}.
% The model takes multi-view images and  LiDAR point clouds as input and 
The proposed~\methodName{} is equipped with 2 cascaded diffusion decoder layers. 
We set the number of modes~$K_\mathrm{train} = 20$ for training and $K_\mathrm{infer} = 200$ for testing.
The model is trained from scratch on the \textsc{navtrain} split for 100 epochs using the DDPM~\cite{ho2020denoising}, with a timestep $T$ of 1000. The training is distributed across 8 NVIDIA L20 GPUs, with a total batch size of 512, and is optimized using AdamW. 
% The ranker's training leverages a fixed trajectory vocabulary and the output from our frozen, pre-trained diffusion model to learn a scoring function. 
In inference, we use DDIM~\cite{songdenoising} to sample the predictions with only 2 denoising steps.
The resulting candidates are then evaluated by the trained ranker, which selects the highest-scoring trajectory as the output. We predict $T_f=8$ timesteps and the interval between each time step is 0.5s.
For more details, please refer to the supplementary material.
\begin{figure*}[t!]
    \centering
    \includegraphics[width=\linewidth]{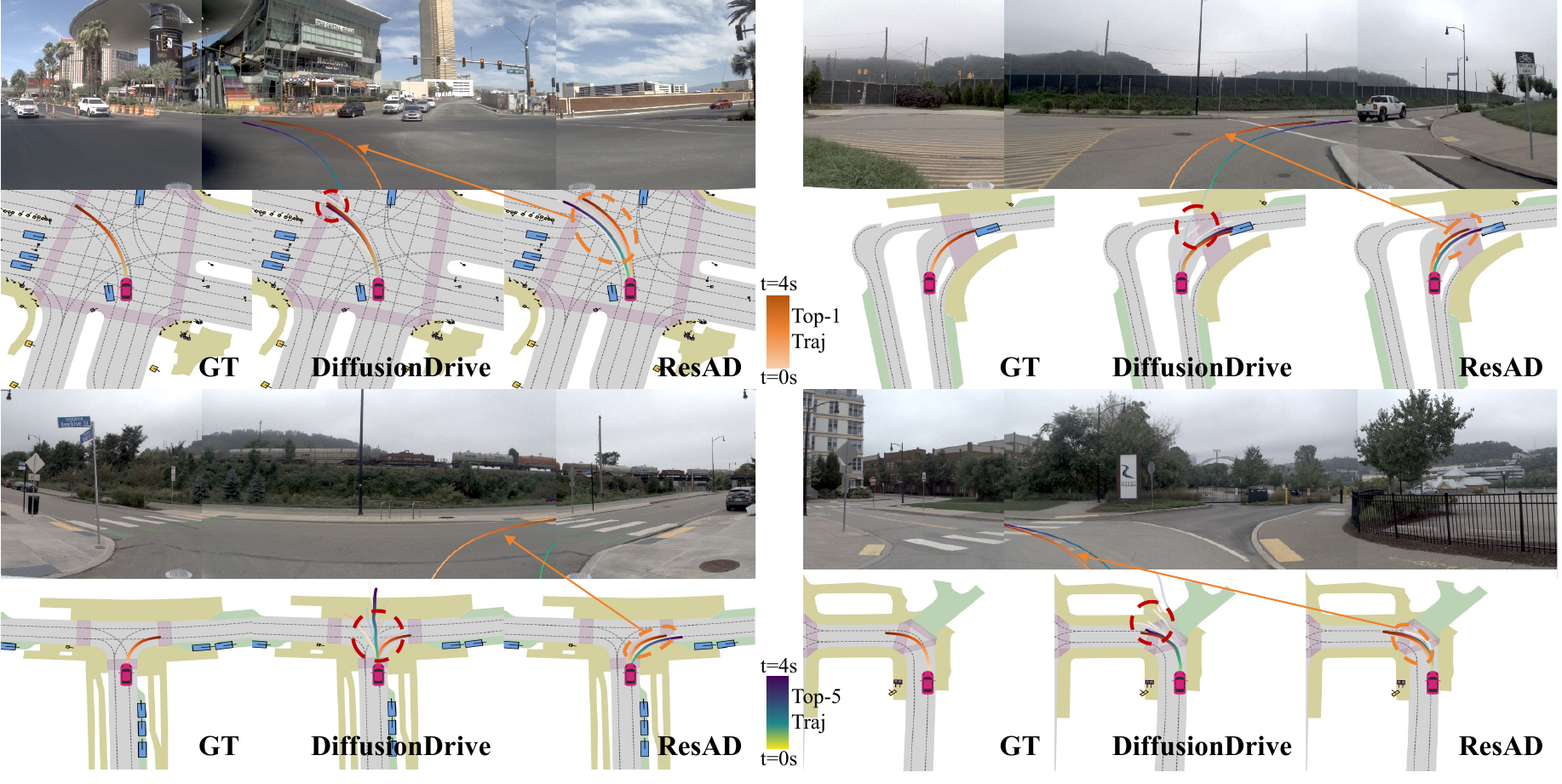}
    \caption{\textbf{Visual comparison of}~\methodName{}. 
    This figure compares the 20 trajectory candidates from~\methodName{} and DiffusionDrive. DiffusionDrive relies on a static, context-agnostic vocabulary, often proposing infeasible or irrelevant trajectories (highlighted by \textcolor[HTML]{BF0500}{red} circles). 
    In contrast,~\methodName{} dynamically generates a set of context-aware trajectories via IRP. This demonstrates our method's more efficient multimodal exploration, which avoids wasting capacity on invalid options and leads to a higher multimodal planning quality ($\mathcal{P}_{m}$), as validated in Sec.~\ref{sec:abl}.
    }
    % The figure shows outputs of 20 candidates from and DiffusionDrive. We highlight the Top-1 and the 5th-ranked trajectory.
    % DiffusionDrive relies on a static, context-agnostic vocabulary, often proposing infeasible trajectories (circled in red). In contrast, the proposed \methodName{} dynamically generates context-aware trajectories by perturbing the ego-vehicle's velocity, addressing limitations in static approaches.}
    \label{fig:vis}
\end{figure*}

\subsection{Main Results}
\boldparagraph{Quantitative Comparison.}
The results presented in Tab.~\ref{tab:navsim_v1} show that~\methodName{} achieves a SOTA performance on the NAVSIM v1 \textsc{navtest} split, with a PDMS of 88.8 using a ResNet-34 backbone, surpassing strong baselines such as DiffusionDrive (PDMS 88.1) and WoTE (PDMS 88.3) under the same backbone. 
With a stronger V2-99 backbone, \methodName{} further improves to PDMS 90.6, surpassing GoalFlow and Hydra-MDP. 
On the more challenging NAVSIM v2 benchmark, the advantages of~\methodName{} are further extended. As shown in Tab.~\ref{tab:navsimv2},~\methodName{} reaches the best or second-best performance across almost all extended sub-metrics. 
Relative to DiffusionDrive, ~\methodName{} improves notably on EP (88.2 vs. 87.5) and DAC (97.2 vs. 95.9), indicating stronger route completion and adherence to drivable areas under the richer EPDMS protocol.

\boldparagraph{Qualitative Comparison.}
A qualitative comparison on NAVSIM (Fig.~\ref{fig:vis}) highlights the different multimodal strategies of \methodName{} and DiffusionDrive. While both successfully avoid the mode collapse typical of vanilla diffusion, their underlying approaches diverge significantly. 
DiffusionDrive relies on a static, predefined trajectory vocabulary. This context-agnostic approach forces it to generate many irrelevant or unfeasible options, such as proceeding straight in a sharp turn scenario (\textbf{highlighted by red circles in the figure}). Although a subsequent filtering step can prune these invalid paths, this two-stage process is inherently inefficient.
In contrast, \methodName{} overcomes this limitation through its distinct trajectory modeling shift. 
This is achieved through a mechanism of perturbing the ego-vehicle's velocity. It directly explores a set of plausible behaviors, generating trajectories that are inherently consistent with the immediate driving context.
A \textbf{real-world vehicle demonstration} of the proposed method is available in the supplementary material.
\subsection{Ablation Studies and Analysis}
\label{sec:abl}
\begin{table}[!b] 
    \centering % 表格居中
    \caption{\textbf{Ablation study on the influence of each component.}}
    \label{tab:abl1} 
    \resizebox{\linewidth}{!}{
        \begin{tabular}{llcccccc}
            \toprule
            {Model} & {Description} & $\textbf{NC}\uparrow$ & $ \textbf{DAC} \uparrow$ & $\textbf{EP} \uparrow$ & $\textbf{TTC} \uparrow$ & $\textbf{C} \uparrow$ & $\textbf{PDMS} \uparrow$ \\
            \midrule
            $\mathcal{M}_0$ & Base Model & 97.8 & 94.2 & 78.1 & 93.4 & 100 & 84.9 \\
            $\mathcal{M}_1$ & $\mathcal{M}_0$ + Ranker & 98.3 & 94.3 & 77.8 & 94.6 & 100 & 85.1 \\
            $\mathcal{M}_2$ & $\mathcal{M}_1$ + TRM & 97.4 &96.6  &80.3  &  93.2& 100 & 86.3 \\
            $\mathcal{M}_3$ & $\mathcal{M}_2$ + PRNorm  & 97.6  &96.7  &81.4  & 93.3 & 100 & 87.2  \\
            $\mathcal{M}_4$ & $\mathcal{M}_3$ + IRP  & 98.0 & 97.5 & 83.4 & 94.1 & 100 & 88.8 \\
            \bottomrule
        \end{tabular}%
    }
\end{table}
\begin{figure}[!t]
    \centering
    \includegraphics[width=\columnwidth]{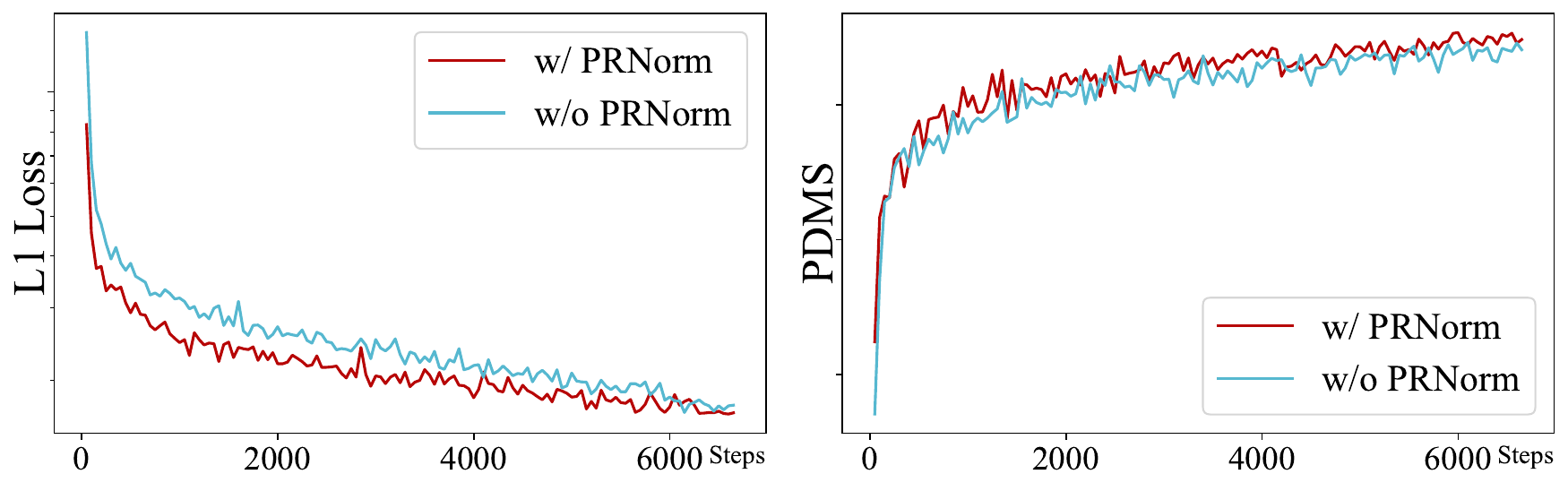}
    \caption{\textbf{PRNorm stabilizes optimization and accelerates convergence.} This figure compares the training dynamics of~\methodName{} with and without PRNorm.
    % This figure shows the loss and mean PDMS curves for ~\methodName{} trained with or without the proposed PRNorm.
    % The L1 loss is calculated between the predicted trajectory and the GT.
    % The model equipped with PRNorm exhibits faster convergence and a lower overall loss.
    % The impact of PRNorm on training efficiency and performance. PRNorm not only accelerates model convergence, evidenced by the significantly faster decline in the loss curve compared to the baseline, but also achieves a consistently higher PDMS score throughout the training process.
    }
    \label{fig:loss}
\end{figure}

\begin{figure}[t!]
    \centering
    \includegraphics[width=\linewidth]{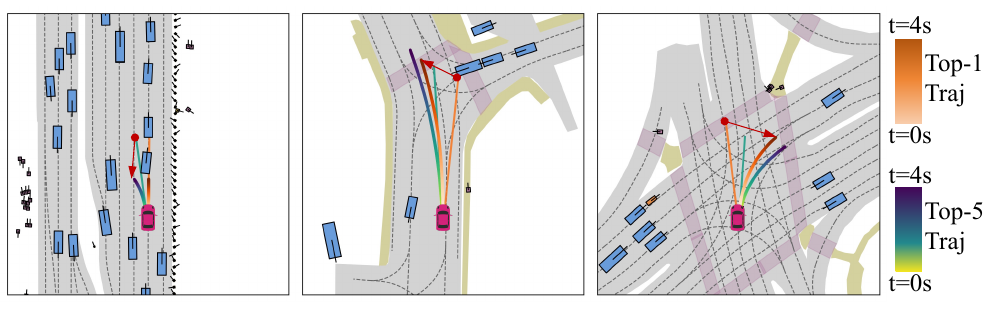}
    \caption{\textbf{The impact of Inertial Reference.} We highlight the 1st-ranked and 5th-ranked trajectory with their corresponding IRs.~\textcolor[HTML]{BF0500}{Red} arrows are used to highlight key examples of the predicted residual.
    }
    \label{fig:ir}
\end{figure}

\boldparagraph{Component Analysis.}
As summarized in Tab.~\ref{tab:abl1}, adding the Ranker to the baseline ($\mathcal{M}_0\rightarrow\mathcal{M}_1$) yields only marginal benefit (PDMS +0.2) and mildly improves safety, indicating its choices concentrate in a narrow mode.
Introducing Trajectory Residual Modeling (TRM) delivers the key boost (DAC +2.3, EP +2.5, PDMS +1.2).
By providing a strong physical prior, TRM compels the model to learn the underlying driving logic rather than relying on the spurious correlations it might learn from raw trajectory data. 
This is further qualitatively validated in Fig.~\ref{fig:ir}.
% capture “must-deviate” cues from obstacles.
Adding Point-wise Residual Normalization (PRNorm) further refines performance and stabilizes convergence, improving EP to 81.4 and PDMS to 87.2. PRNorm stabilizes optimization by resolving spatio-temporal scale variance, preventing far-field errors from dominating. This refines the trajectory quality, leading to another notable increase in EP (+1.1).
% By normalizing residuals point-wise, it prevents large-magnitude errors from distant, uncertain waypoints from dominating the optimization.
Finally, Inertial Reference Perturbation (IRP) delivers the largest single improvement, reaching 88.8 PDMS (+1.6 over $\mathcal{M}_3$). IRP unlocks true, context-aware multimodality. It generates a diverse yet plausible set of intent hypotheses, providing high-quality candidates for the ranker to evaluate. This is the key that enables the model to excel in complex scenarios, dramatically improving EP (+2.0) and DAC (+0.8).

\boldparagraph{\textbf{Impact of PRNorm on Training Dynamics.}}
As shown in Fig.~\ref{fig:loss}, PRNorm enables a significantly faster decline in loss compared to the baseline (vanilla min-max normalization), accelerating model convergence. 
Furthermore, we calculate the PDMS of the predicted trajectory every step, which is also higher with PRNorm. It demonstrates its comprehensive benefits to both training efficiency and the final performance.

\boldparagraph{\textbf{Qualitative Analysis of the TRM.}}
As shown in Fig.~\ref{fig:ir}, we remodel the trajectory as a residual against the inertial reference (IR). 
The IR can serve as a \textbf{counterfactual baseline}: in steady cruising, predictions largely coincide with the IR, yielding small residuals and stable optimization; in dynamic scenes (merging, intersections, sharp turns, braking), the baseline becomes highly inaccurate, producing strong, directional residuals. These residual vectors point away from obstacles or follow road curvature, indicating the model learns why to deviate (due to obstacles, yielding, or turning) rather than merely fitting a complex trajectory. 
Moreover, IRP samples a bundle of lightly perturbed baselines around the original IR, expanding the physically plausible neighborhood. 
The model learns sharper residuals over these references, avoiding mode collapse. Together, IR + IRP focus capacity on what must change and encourage exploration of a more effective trajectory space, enabling robust predictions with few denoising steps.

% This residual formulation focuses learning on what must change, mitigates spatio-temporal imbalance, and enables efficient prediction with few denoising steps.

% The inertial reference can serve as a counterfactual baseline. In steady cruising, predictions largely coincide with the IR, yielding small residuals and stable optimization; in dynamic scenes (merging, intersections, sharp turns, braking), the baseline becomes highly inaccurate, producing strong, directional residuals that explicitly encode the required deviation induced by obstacles or road curvature. This residual formulation focuses learning on what must change, mitigating spatio-temporal imbalance and enabling efficient prediction.

% 这里应该突出残差建模的有效性.

\begin{table}[t!] 
    \centering 
    \caption{\textbf{Extending Normalized Residual Trajectory Modeling to other models.}} 
    \label{tab:abl2} 
    \resizebox{\linewidth}{!}{
        \begin{tabular}{llcccccc}
            \toprule
            {Model}  & $\textbf{NC}\uparrow$ & $ \textbf{DAC} \uparrow$ & $\textbf{EP} \uparrow$ & $\textbf{TTC} \uparrow$ & $\textbf{C} \uparrow$ & $\textbf{PDMS} \uparrow$ \\
            \midrule
            Transfuser & 97.7 & 92.8 & 79.2 & 92.8 & 100 & 84.0 \\
            \rowcolor{gray!20}
            +TRM & 97.7 & 93.5 & 80.0 & 93.5 & 100 & 85.2 \\
            \rowcolor{gray!20}
            +PRNorm & 98.0 & 94.2 & 79.8 & 93.6 & 99.9 & 85.6 \\
        \midrule
            Transfuser$_\mathrm{DP}$ & 97.4 & 93.5 & 79.0 & 93.0 & 100 & 84.5 \\
            \rowcolor{gray!20}
            +TRM & 98.0 & 93.9 & 80.2 & 93.6 & 100 & 85.5 \\
            \rowcolor{gray!20}
            +PRNorm & 98.2 & 94.8 & 79.4 & 94.2 & 100 & 85.8 \\ 
            \bottomrule
        \end{tabular}%
    }
\end{table}

\begin{table}[!b]
\centering
\caption{\textbf{Comparison of running time and performance.}
FPS and total runtime are measured on an NVIDIA 4090 GPU.}

\resizebox{\linewidth}{!}{
\begin{tabular}{lcccccccc} 
\toprule
Method & PDMS$\uparrow$ & ${K}_\text{infer}$ & $\mathcal{P}_m$ & Total$\downarrow$ & Para.$\downarrow$ & FPS $\uparrow$ \\
\midrule
% Transfuser & 84.0  & - & - & \textbf{0.2ms} & \textbf{56M} & \textbf{60} \\
Transfuser$_\mathrm{DP}$ & 84.6 & 20 & \underline{84.4} & 130.0ms & 101M & 7 \\
DiffusionDrive & \underline{88.1}  & 20 & 60.3 & \textbf{7.6ms} & \textbf{60M} & \textbf{45} \\
\methodName{} w/o Ranker & - & 20 & \textbf{86.1}  & \underline{7.7ms} & \underline{62M} & \textbf{45} \\
\methodName{} & \textbf{88.3} & 20 & \textbf{86.1} & {11.4ms} & {68M} & \underline{37} \\
\bottomrule
\end{tabular}
}
\label{tab:fps}
\end{table}

\boldparagraph{\textbf{Superiority of the Normalized Residual Trajectory Modeling.}}
% 我们通过许多实验证明了Normalized Residual Trajectory Modeling存在的诸多优势：
% (1) 即插即用，可以轻松引入到不同的规划网络中,并带来性能提升.如表tab:abl2
% (2) 能够生成高质量/上下文有关的多模态轨迹.如可视化图3,这里加一个表.
% (3) 更出色的test-time scaling 性能.
We demonstrate the numerous advantages of Normalized Residual Trajectory Modeling (NRTM) through further experiments:
\textbf{(1)~Plug-and-Play.}
NRTM is a drop-in strategy that requires limited changes to the architecture or loss design: attach TRM (+PRNorm) to \emph{any} planner and train as usual. 
As shown in Tab.~\ref{tab:abl2}, it consistently lifts performance on both an MLP planner (Transfuser) and a diffusion planner (Transfuser$_\mathrm{DP}$). 
On Transfuser, TRM raises PDMS from 84.0 to 85.2, and TRM+PRNorm further to 85.6; comparable gains are observed on Transfuser$_\mathrm{DP}$. 
By residual modeling, NRTM captures essential high-order driving primitives and, via spatio-temporal data alignment, accelerates optimization. It delivers method-agnostic improvements to E2EAD safety and efficiency with minimal integration cost.
\textbf{(2)~Efficient and Context-aware Multimodal Planning.}
As shown in Tab.~\ref{tab:fps}, compared with DiffusionDrive, all increases in parameters and inference time for~\methodName{} come from the Ranker, which selects the optimal trajectory from a set of candidates. Beyond reporting PDMS, we also introduce the mean PDMS $\mathcal{P}_m$
to evaluate the average quality of the model’s multimodal trajectory outputs.~\methodName{} attains a substantially higher $\mathcal{P}_m$, providing further quantitative evidence that~\methodName{} generates context-aware multimodal trajectories.
\textbf{(3)~Test-Time Scaling.}
\begin{figure}[t!]
    \centering
    \includegraphics[width=\linewidth]{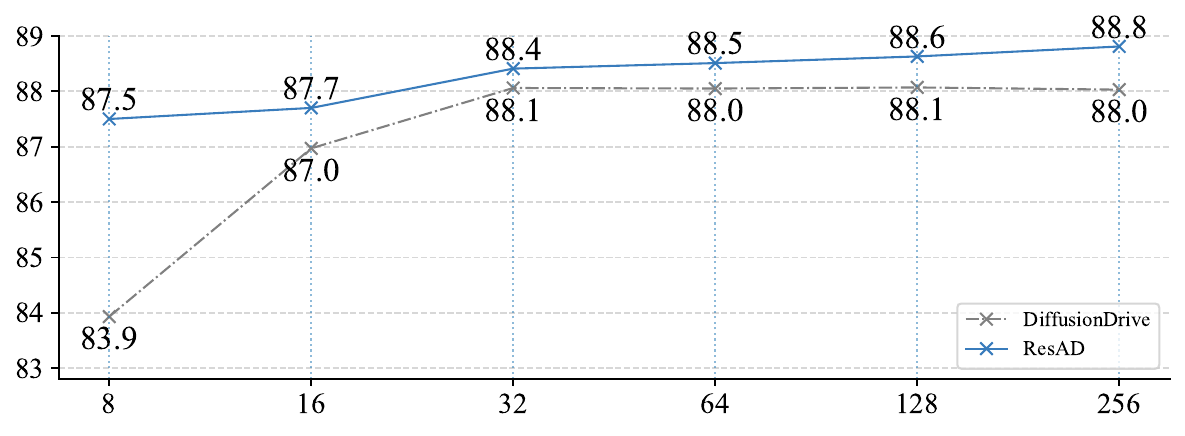}
    \caption{\textbf{Dynamic number of ${K}_\text{infer}$}. This figure plots PDMS as a function of the inference-time sampled noise numbers ${K}_\text{infer}$.
    }
    \label{fig:tts}
\end{figure}
As shown in Fig.~\ref{fig:tts}, while DiffusionDrive exhibits a distinct performance drop when $K_\text{infer} < K_\text{train}$ and remains around 88.1 when $K_\text{infer} > 20$, ~\methodName{} realizes a consistent performance and continuous growth.
% significantly outperforming the baseline.
The introduction of IR provides effective physical prior, constraining predictions within a context-relevant region and effectively avoiding invalid samples. 
Therefore, NRTM not only boosts the performance but also converts extra inference compute into consistent PDMS gains, revealing a compute-adaptive, anytime advantage at deployment.

\section{Conclusion}
In this work, we revisit the conventional E2EAD paradigm, highlighting a key challenge where directly predicting raw trajectories can create a severe optimization imbalance and risks learning spurious correlations.
We propose~\methodName{}, a novel framework that fundamentally reframes the learning problem. Our approach introduces two key innovations: Trajectory Residual Modeling (TRM) and Point-wise Residual Normalization (PRNorm).
TRM provides a strong physical prior via an inertial reference, forcing the model to learn only the necessary, context-driven residuals. This simplifies the task and focuses capacity on learning robust driving logic, not spurious patterns. PRNorm mitigates the optimization imbalance, ensuring that safety-critical, near-term adjustments are prioritized over uncertain long-term predictions.
Our state-of-the-art results on NAVSIM are a direct consequence of this conceptual shift. By taming the imbalanced data and simplifying the learning objective,~\methodName{} provides a significantly more robust, stable, and scalable foundation for E2EAD systems.

% \input{doc/intro}
% \input{doc/related}
% \input{doc/algo}
% \input{doc/exp}
% \input{doc/limit_and_conclu}
% \section*{Acknowledgement}
% We would like to acknowledge Qingjie Wang, Yongjun Yu, Zehua Li, Peng Wang, Nuoya Zhou, Songlin Yang, Ruiqi Wang, Tianheng Cheng, Changze Li, Zhe Chen, and Tong Qin for discussion and assistance.


{
    \small
    \bibliographystyle{ieeenat_fullname}
    \bibliography{main}
}

% \appendix
% \beginsupplement
% \input{content/appendix}

\end{document}